# Automatically Training a Problematic Dialogue Predictor for a Spoken Dialogue System

**Marilyn A. Walker** WALKER@RESEARCH.ATT.COM
**Irene Langkilde-Geary** ILANGKIL@ISI.EDU
**Helen Wright Hastie** HHASTIE@RESEARCH.ATT.COM
**Jerry Wright** JWRIGHT@RESEARCH.ATT.COM
**Allen Gorin** ALGOR@RESEARCH.ATT.COM
*AT&T Shannon Laboratory*
*180 Park Ave., Bldg 103, Room E103*
*Florham Park, NJ 07932*

## Abstract

Spoken dialogue systems promise efficient and natural access to a large variety of information sources and services from any phone. However, current spoken dialogue systems are deficient in their strategies for preventing, identifying and repairing problems that arise in the conversation. This paper reports results on automatically training a Problematic Dialogue Predictor to predict problematic human-computer dialogues using a corpus of 4692 dialogues collected with the *How May I Help You*$^{SM}$ spoken dialogue system. The Problematic Dialogue Predictor can be immediately applied to the system's decision of whether to transfer the call to a human customer care agent, or be used as a cue to the system's dialogue manager to modify its behavior to repair problems, and even perhaps, to prevent them. We show that a Problematic Dialogue Predictor using automatically-obtainable features from the first two exchanges in the dialogue can predict problematic dialogues 13.2% more accurately than the baseline.

## 1. Introduction

Spoken dialogue systems promise efficient and natural access to a large variety of information sources and services from any phone. Systems that support short utterances to select a particular function (through a statement such as "Say credit card, collect or person-to-person") are saving companies millions of dollars per year. Deployed systems and research prototypes exist for applications such as personal email and calendars, travel and restaurant information, and personal banking (Baggia, Castagneri, & Danieli, 1998; Walker, Fromer, & Narayanan, 1998; Seneff, Zue, Polifroni, Pao, Hetherington, Goddeau, & Glass, 1995; Sanderman, Sturm, den Os, Boves, & Cremers, 1998; Chu-Carroll & Carpenter, 1999) *inter alia*. Yet there are still many research challenges: current systems are limited in the interaction they support and brittle in many respects.

This paper investigates methods by which spoken dialogue systems can *learn* to support more natural interaction on the basis of their previous experience. One way that current spoken dialogue systems are quite limited is in their strategies for detecting and repairing problems that arise in conversation, such as misunderstandings due to speech recognition error or misinterpretation. If a problem can be detected, the system can either transfer the call to a human customer care agent or modify its dialogue strategy in an attempt to repair the problem. We can train systems to improve their ability to detect problems by exploiting dialogues collected in interactions with human users where the initial segments of these dialogues are used to train a Problematic Dialogue Predictor (PDP) to *predict* that a problem is likely to occur. The output of the PDP can be immediately applied to the system's decision of whether to transfer the call to a human customer care agent, or it could potentially be used as a cue to the system's Dialogue Manager to modify its behavior to repair problems, and even perhaps, to prevent them.

In previous work, we reported initial results for training a PDP using a variety of different feature sets (Langkilde, Walker, Wright, Gorin, & Litman, 1999; Walker, Langkilde, Wright, Gorin, & Litman, 2000b). When analyzing the performance of the fully automatic feature set, we examined which hand-labelled features made large performance improvements, under the assumption that future work should focus on developing automatic features that approximate the information provided by these hand-labelled features. The analysis indicated that the hand-labelled *SLU-success* feature, which encodes whether the spoken language understanding (SLU) component captured the meaning of each exchange correctly. When this hand-labelled feature is added to the automatic features, it improved the performance of the PDP by almost 7.6%. This finding led us to develop an *SLU-success* predictor (Walker, Wright, & Langkilde, 2000c) and a new version of the PDP that we report on here. The new version of the PDP takes as input a fully automatic version of the *SLU-success* feature, which we call *auto-SLU-success*.

We train and test both the *auto-SLU-success* predictor and the PDP on a corpus of 4692 dialogues collected in an experimental trial of AT&T's *How May I Help You* (HMIHY$^{SM}$) spoken dialogue system (Gorin, Riccardi, & Wright, 1997; Abella & Gorin, 1999; Riccardi & Gorin, 2000; E. Ammicht & Alonso, 1999). In this trial, the HMIHY system was installed at an AT&T customer care center. HMIHY answered calls from live customer traffic and successfully automated a large number of customer requests. An example dialogue that HMIHY completed successfully is shown in Figure 1. The phone numbers, card numbers, and pin numbers in the sample dialogues are artificial.

| | |
|---|---|
| S1: | AT&T How may I help you? |
| U1: | I need to [ uh ] put a call on my calling card please |
| S2: | May I have your card number, please? |
| U2: | 7 6 5 4 3 2 1 0 9 8 7 6 5 4 |
| S3: | What number would you like to call? |
| U3: | 8 1 4 7 7 7 6 6 6 6 (*misunderstood*) |
| S4: | May I have that number again? |
| U4: | 8 1 4 7 7 7 6 6 6 6 |
| S5: | Thank you. |

Figure 1: Sample TASKSUCCESS Dialogue

Note that the system's utterance in S4 consists of a repair initiation, motivated by the system's ability to detect that the user's utterance U3 was likely to have been misunderstood. The goal of the *auto-SLU-success* predictor is to improve the system's ability to detect such misunderstandings. The dialogues that have the desired outcome, in which HMIHY successfully automates the customer's call, are referred to as the TASKSUCCESS dia-logues. Dialogues in which the HMIHY system did not successfully complete the caller's task are referred to as PROBLEMATIC. These are described in further detail below.

This paper reports results from experiments that test whether it is possible to learn to automatically predict that a dialogue will be problematic on the basis of information the system has: (1) early in the dialogue; and (2) in real time. We train an automatic classifier for predicting problematic dialogues from features that can be automatically extracted from the HMIHY corpus. As described above, one of these features is the output of the *auto-SLU-success* predictor, the *auto-SLU-success* feature, which predicts whether or not the current utterance was correctly understood (Walker et al., 2000c). The results show that it is possible to *predict* problematic dialogues using fully automatic features with an accuracy ranging from 69.6% to 80.3%, depending on whether the system has seen one or two exchanges. It is possible to *identify* problematic dialogues with an accuracy up to 87%.

Section 2 describes HMIHY and the dialogue corpus that the experiments are based on. Section 3 discusses the type of machine learning algorithm adopted, namely RIPPER and gives a description of the experimental design. Section 4 gives a breakdown of the features used in these experiments. Section 5 presents the method of predicting the feature *auto-SLU-success* and gives accuracy results. Section 6 presents methods used for utilizing RIPPER to train the automatic Problematic Dialogue Predictor and gives the results. We delay our discussion of related work to Section 7 when we can compare it to our approach. Section 8 summarizes the paper and describes future work.

## 2. The HMIHY Data

HMIHY is a spoken dialogue system based on the notion of *call routing* (Gorin et al., 1997; Chu-Carroll & Carpenter, 1999). In the HMIHY call routing system, services that the user can access are classified into 14 categories, plus a category called *other* for tasks that are not covered by the automated system and must be transferred to a human operator (Gorin et al., 1997). Each category describes a different task, such as person-to-person dialing, or receiving credit for a misdialed number. The system determines which task the caller is requesting on the basis of its understanding of the caller's response to the open-ended system greeting *AT&T, How May I Help You?*. Once the task has been determined, the information needed for completing the caller's request is obtained using dialogue submodules that are specific for each task (Abella & Gorin, 1999).

The HMIHY system consists of an automatic speech recognizer, a spoken language understanding module, a dialogue manager, and a computer telephony platform. During the trial, the behaviors of all the system modules were automatically recorded in a log file, and later the dialogues were transcribed by humans and labelled with one or more of the 15 task categories, representing the task that the caller was asking HMIHY to perform, on a per utterance basis. The log files also included labels indicating whether the wizard had taken over the call or the user had hung up. Our experiments use the log files to extract automatically obtainable features used as predictors, and to define the classes of dialogues that we want to learn to predict. The corpus of 4692 dialogues used in our experiments was collected in several experimental trials of HMIHY on live customer traffic (Riccardi & Gorin, 2000; E. Ammicht & Alonso, 1999), and is referred to as HM2 in (Riccardi & Gorin, 2000).

The dialogues vary in length, 97% are five exchanges or less with 23% of all the dialogues consisting of only two exchanges.

As mentioned above, dialogues in which HMIHY successfully automates the customer's call, as illustrated in Figure 1, are referred to as TASKSUCCESS. Other calls, which are problematic, are divided into three categories. The first category, referred to as HANGUP, results from a customer's decision to hang up on the system. A sample HANGUP dialogue is in Figure 2. A caller may hang up because s/he is frustrated with the system; our goal is to learn from the corpus which system behaviors led to the caller's frustration.

| | |
|---|---|
| S1: | AT&T How may I help you? |
| U1: | I need credit please. |
| S2: | Okay. What is the reason for the credit? |
| U2: | Miss, uh, Miss, different area code than I needed. |
| S3: | Sorry. I need to know whether you need credit for a wrong number, bad connection or a call that was cut off. |
| U3: | It is a wrong number. (*impatient humming*) |
| S4: | What was the number that you dialed? |
| | USER HANGS UP |

Figure 2: Sample HANGUP Dialogue

The second problematic category (WIZARD), results from a human customer care agent's decision to take over the call from the system. Because HMIHY is experimental, each call during the field trial was monitored by a human agent serving as a wizard who could override the system. There were a number of agents who participated as wizards during the trial of HMIHY and each wizard was simply told to take over the call if s/he perceived problems with the system's performance. The wizard's decision was logged by the experimental setup, resulting in labelling the call as one that the wizard took over. Of course we can only infer what might have motivated the wizard to take over the call, but we assume that the wizard had good reason for doing so. A dialogue where the wizard decided that the dialogue was problematic and took over the call is shown in Figure 3.

| | |
|---|---|
| S1: | AT&T How may I help you? |
| U1: | (*silence*) |
| S2: | Sorry. Please briefly tell me how I may help you? |
| U2: | I'm trying to call 1 8 hundred call A T T. |
| S3: | Do you want to place a call? |
| U3: | Through my calling card. |
| S4: | May I have your card number, please? |
| U4: | It's 8 7 6 5 4 3 2 1 0 and then my pin number is 8 7 6 5. (*misrecognized*) |
| S5: | Sorry. Please enter or say your card number again. |
| | WIZARD STEPS IN |

Figure 3: Sample WIZARD Dialogue

The third problematic category, the TASKFAILURE dialogues, are cases where the system completed the call, but carried out a task that was not the one that the customer was actually requesting. An example TASKFAILURE dialogue is given in Figure 4: HMIHY interpreted utterance U2 as a request to make a third-party call e.g. to *bill it to my home phone*. HMIHY then asked the caller for the information it needed to carry out this task, the caller complied, and the system completed the call.

| | |
|---|---|
| S1: | AT&T How may I help you? |
| U1: | (*silence*) |
| S2: | Sorry. Please briefly tell me how I may help you? |
| U2: | I wanta find out how much my bill's worth. |
| S3: | What phone number would you like the call billed to? |
| U3: | *Um* the access number. |
| S4: | Sorry. Could I have that phone number again? |
| U4: | 1 8 hundred 2 2 5 5 2 8 8. |
| S5: | What number would you like to call? |
| U5: | 800 225 5288 (*touchtone input*) |
| S6: | Thank you. |

Figure 4: Sample TASKFAILURE Dialogue

## 3. Experimental Design

Our experiments apply the machine learning program RIPPER (Cohen, 1995, 1996) to automatically classify the dialogues as problematic or successful. RIPPER is a fast and efficient rule learning system described in more detail in (Cohen, 1995, 1996); we describe it briefly here for completeness. RIPPER is based on the *incremental reduced error pruning* (IREP) algorithm described in (Furnkranz & Widmer, 1994). RIPPER improves on IREP with an information gain metric to guide rule pruning and a Minimum Description Length or MDL-based heuristic for determining how many rules should be learned (see Cohen 1995, 1996 for more details). Like other learners, RIPPER takes as input the names of a set of *classes* to be learned, the names and ranges of values of a fixed set of *features*, and *training data* specifying the class and feature values for each example in a training set. Its output is a *classification model* for predicting the class of future examples, expressed as an ordered set of if-then rules.

Although any one of a number of learners could be applied to this problem, we had a number of reasons for choosing RIPPER. First, it was important to be able to integrate the results of applying the learner back into the HMIHY spoken dialogue system. Previous work suggests that the if-then rules that RIPPER uses to express the learned classification model are easy for people to understand (Catlett, 1991; Cohen, 1995), making it easier to integrate the learned rules into the HMIHY system. Second, RIPPER supports continuous, symbolic and textual bag (set) features (Cohen, 1996), while other learners, such as Classification and Regression trees (CART) (Brieman, Friedman, Olshen, & Stone, 1984), do not support textual bag features. There are several textual features in this dataset that prove useful in classifying the dialogues. One of the features that we wished to use was the string representing the recognizer's hypothesis. This is supported in RIPPER because there is no *a priori* limitation on the size of the set. The usefulness of the textual features is exemplified in Section 6.3. Finally, previous work in which we had applied other learners to the *auto-SLU-* *success* predictor, utilizing the best performing feature set with the textual bag features removed, suggested that we could not expect any significant performance improvements from using other learners (Walker et al., 2000c).

In order to train the problematic dialogue predictor (PDP), RIPPER uses a set of features. As discussed above, initial experiments showed that the hand-labelled *SLU-success* feature, which encodes whether an utterance has been misunderstood or not, is highly discriminatory in identifying problematic dialogues. However, all the features used to train the PDP must be totally automatic if we are to use the PDP in a working spoken dialogue system. In order to improve the performance of the fully automatic PDP, we developed a fully automatic approximation of the hand-labelled feature, which we call the *auto-SLU-success* feature, in separate experiments with RIPPER. The training of the *auto-SLU-success* feature is discussed in Section 5.

Evidence from previous trials of HMIHY suggest that it is important to identify problems within a couple of exchanges and 97% of the dialogues in the corpus are five exchanges or less. Thus features for the first two exchanges are encoded since the goal is to *predict* failures before they happen. The experimental architecture of the PDP is illustrated in Figure 5. This shows how RIPPER is used first to predict *auto-SLU-success* for the first and second exchanges. This feature is fed into the PDP along with the other automatic features. The output of the PDP determines whether the system continues, or if a problem is predicted, the Dialogue Manager may adapt its dialogue strategy or transfer the customer to a customer agent.

Since 23% of the dialogues consisted of only two exchanges, we exclude the second exchange features for those dialogues where the second exchange consists only of the system playing a closing prompt. We also excluded any features that indicated to the classifier that the second exchange was the last exchange in the dialogue. We compare results for *predicting* problematic dialogues, with results for *identifying* problematic dialogues, when the classifier has access to features representing the whole dialogue.

In order to test the *auto-SLU-success* predictor as input to the PDP, we first defined a training and test set for the combined problem. The test set for the *auto-SLU-success* predictor contains the exchanges that occur in the dialogues of the PDP test set. We selected a random 867 dialogues as the test set and then extracted the corresponding exchanges (3829 exchanges). Similarly for training, the PDP training set contains 3825 dialogues which corresponds to a total of 16901 exchanges for training the *auto-SLU-success* predictor.

The feature *auto-SLU-success* is predicted for each utterance in the test set, thus enabling the system to be used on new data without the need for hand-labelling. However, there are two possibilities for the origin of this feature in the training set. The first possibility is for the training set to also consist of solely automatic features. This method has the potential advantage that the trained PDP will compensate, if necessary, for whatever noise exists in the *auto-SLU-success* predictions (Wright, 2000). An alternative to training the PDP on the automatically derived *auto-SLU-success* feature is to train it on the hand-labelled *SLU-success* while still testing it on the automatic feature. This second method is referred to as "hand-labelled-training" or *hlt-SLU-success*. This may provide a more accurate model but it may not capture the characteristics of the automatic feature in the test set. Results for these two methods are presented in Section 6.4.

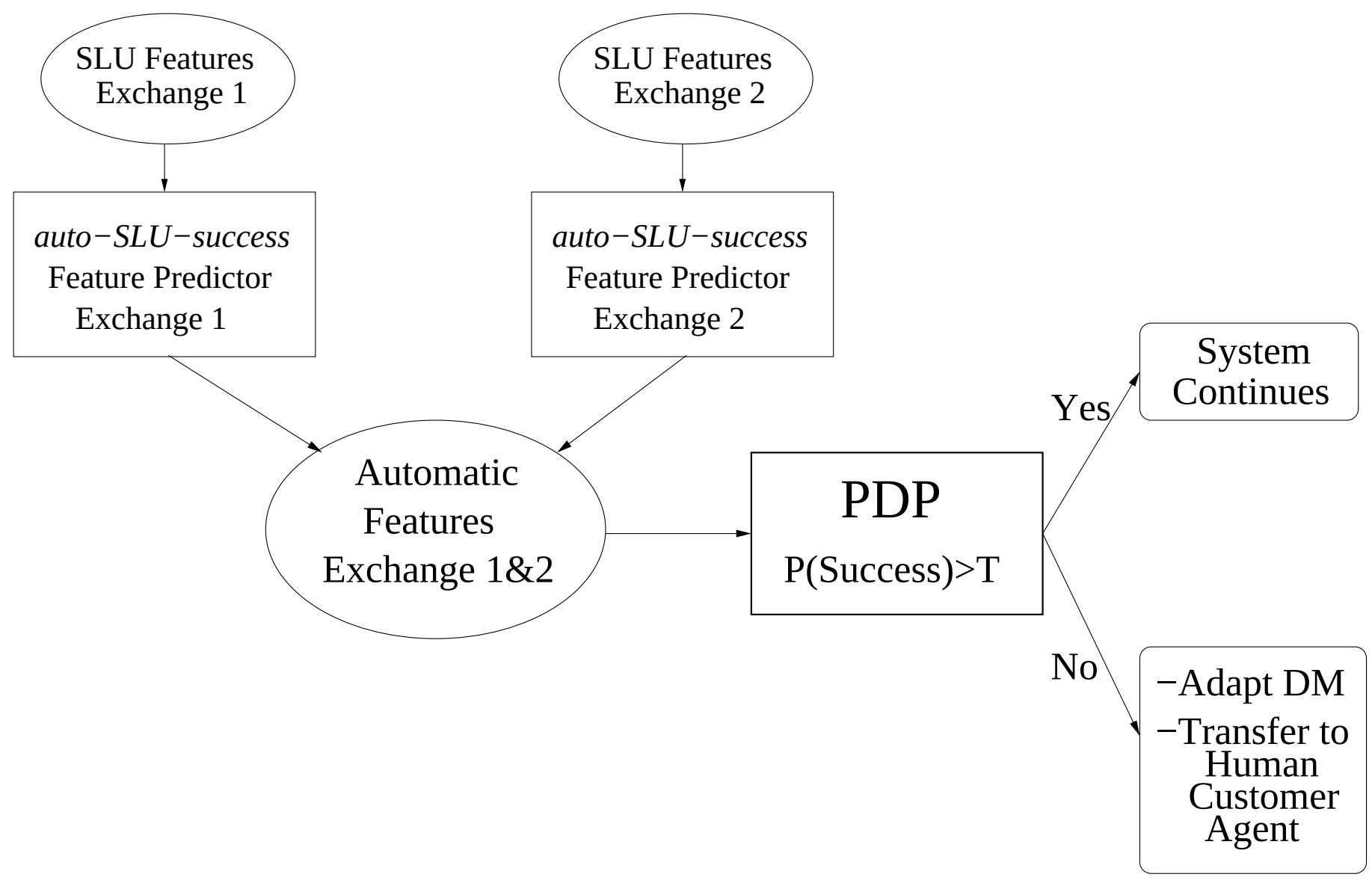

Figure 5: System architecture using features from the first 2 exchanges

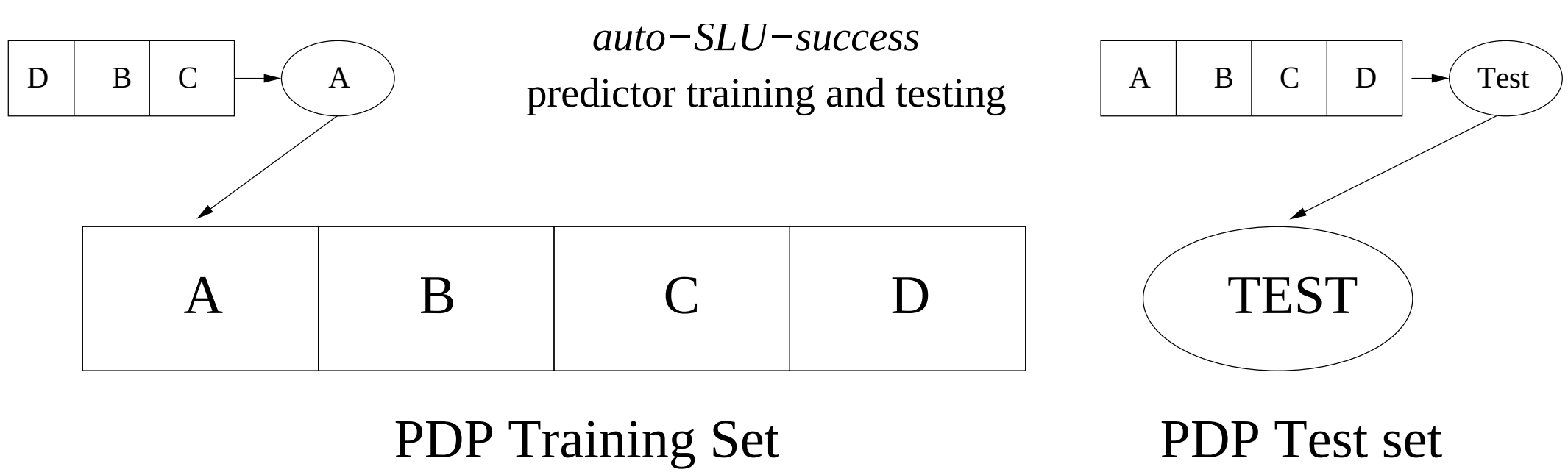

Figure 6: Data for segmentation using cross-validation

The problem with using *auto-SLU-success* for training the PDP is that the same data is used to train the *auto-SLU-success* predictor. Therefore, we used a cross-validation technique (also known as jack-knifing) (Weiss & Kulikowski, 1991), whereby the training set is partitioned into 4 sets. Three of these sets are used for training and the fourth for testing. The results for the fourth set are noted and the process is repeated, rotating the sets from training to testing. This results in a complete list of predicted *auto-SLU-success* for the training set. The features for the test set exchanges are derived by training RIPPER on the whole training set. This process is illustrated in Figure 6.

The following section gives a breakdown of the input features. Section 5 describes the training and results of the *auto-SLU-success* predictor and Section 6 reports the accuracy results for the PDP.

## 4. The Features

- **Acoustic/ASR Features**
  - recog, recog-numwords, asr-duration, dtmf-flag, rg-modality, rg-grammar, tempo
- **SLU Features**
  - a confidence measure for each of the 15 possible tasks that the user could be trying to do
  - salience-coverage, inconsistency, context-shift, top-task, nexttop-task, top-confidence, diff-confidence, confpertime, salpertime, auto-SLU-success
- **Dialogue Manager and Discourse History Features**
  - sys-label, utt-id, prompt, reprompt, confirmation, subdial
  - running tallies: num-utts, num-reprompts, percent-reprompts, num-confirms, percent-confirms, num-subdials, percent-subdials
  - whole dialogue: dial-duration.
- **Hand-Labelled Features**
  - tscript, human-label, age, gender, user-modality, clean-tscript, cltscript-numwords, SLU-success

Figure 7: Features for spoken dialogues.

A dialogue consists of a sequence of exchanges where each exchange consists of one turn by the system followed by one turn by the user. Each dialogue and exchange is encoded using the set of 53 features in Figure 7. Each feature was either automatically logged by one of the system modules, hand-labelled by humans, or derived from raw features. The hand-labelled features are used to produce a TOPLINE, an estimation of how well a classifier could do that had access to perfect information. To see whether our results can generalize, we also experiment with using a subset of features that are task-independent described in detail below.

Features logged by the system are utilized because they are produced automatically, and thus can be used during runtime to alter the course of the dialogue. The system modules for which logging information was collected were the acoustic processor/automatic speech recognizer (ASR) (Riccardi & Gorin, 2000), the spoken language understanding (SLU) module (Gorin et al., 1997), and the Dialogue Manager (DM) (Abella & Gorin, 1999). Each module and the features obtained from it are described below.

**Automatic Speech Recognition**: The automatic speech recognizer (ASR) takes as input the caller's speech and produces a potentially errorful transcription of what it believes the caller said. The ASR features for each exchange include the output of the speech recognizer (*recog*), the number of words in the recognizer output (*recog-numwords*), the duration in seconds of the input to the recognizer (*asr-duration*), a flag for touchtone in-put (*dtmf-flag*), the input modality expected by the recognizer (*rg-modality*) (one of: none, speech, touchtone, speech+touchtone, touchtone-card, speech+touchtone-card, touchtone-date, speech+touchtone-date, or none-final-prompt), and the grammar used by the recognizer (*rg-grammar*) (Riccardi & Gorin, 2000). We also calculate a feature called *tempo* by dividing the value of the *asr-duration* feature by the *recog-numwords* feature.

The motivation for the ASR features is that any one of them may reflect recognition performance with a concomitant effect on spoken language understanding. For example, other work has found *asr-duration* to be correlated with incorrect recognition (Hirschberg, Litman, & Swerts, 1999). The name of the grammar (*rg-grammar*) could also be a predictor of SLU errors since it is well known that the larger the grammar is, the more likely an ASR error is. In addition, the *rg-grammar* feature also encodes expectations about user utterances at that point in the dialogue, which may correlate to differences in the ease with which any one recognizer could correctly understand the user's response. One motivation for the *tempo* feature is that previous work suggests that users tend to slow down their speech when the system has misunderstood them (Levow, 1998; Shriberg, Wade, & Price, 1992); this strategy actually leads to more errors since the speech recognizer is not trained on this type of speech. The *tempo* feature may also indicate hesitations, pauses, or interruptions, which could also lead to ASR errors. On the other hand, touchtone input in combination with speech, as encoded by the feature *dtmf-flag*, might increase the likelihood of understanding the speech: since the touchtone input is unambiguous it can constrain spoken language understanding.

**Spoken Language Understanding**: The goal of the spoken language understanding (SLU) module is to identify which of the 15 possible tasks the user is attempting and extract from the utterance any items of information that are relevant to completing that task, e.g. a phone number is needed for the task *dial for me*.

Fifteen of the features from the SLU module represent the distribution for each of the 15 possible tasks of the SLU module's confidence in its belief that the user is attempting that task (Gorin et al., 1997). We also include a feature to represent which task has the highest confidence score (*top-task*), and which task has the second highest confidence score (*nexttop-task*), as well as the value of the highest confidence score (*top-confidence*), and the difference in values between the top and next-to-top confidence scores (*diff-confidence*).

Other features represent other aspects of the SLU processing of the utterance. The *inconsistency* feature is an intra-utterance measure of semantic diversity, according to a task model of the domain (Abella & Gorin, 1999). Some task classes occur together quite naturally within a single statement or request, e.g. the *dial for me* task is compatible with the *collect call* task, but is not compatible with the *billing credit* task. The *salience-coverage* feature measures the proportion of the utterance which is covered by the salient grammar fragments. This may include the whole of a phone or card number if it occurs within a fragment. The *context-shift* feature is an inter-utterance measure of the extent of a shift of context away from the current task focus, caused by the appearance of salient phrases that are incompatible with it, according to a task model of the domain.

In addition, similar to the way we calculated the *tempo* feature, we normalize the *salience-coverage* and *top-confidence* features by dividing them by *asr-duration* to produce the *salpertime* and *confpertime* features. The *tempo* and the *confpertime* and *salpertime* features are used only for predicting *auto-SLU-success*.

The motivation for these SLU features is to make use of information that the SLU module has as a result of processing the output of ASR and the current discourse context. For example, for utterances that follow the first utterance, the SLU module knows what task it believes the caller is trying to complete. The *context-shift* feature incorporates this knowledge of the discourse history, with the motivation that if it appears that the caller has changed her mind, then the SLU module may have misunderstood an utterance.

**Dialogue Manager**: The function of the Dialogue Manager is to take as input the output of the SLU module, decide what task the user is trying to accomplish, decide what the system will say next, and update the discourse history (Abella & Gorin, 1999). The Dialogue Manager decides whether it believes there is a single unambiguous task that the user is trying to accomplish, and how to resolve any ambiguity.

Features based on information that the Dialogue Manager logged about its decisions or features representing the ongoing history of the dialogue might be useful predictors of SLU errors or task failure. Some of the potentially interesting Dialogue Manager events arise due to low SLU confidence levels which lead the Dialogue Manager to *reprompt* the user or *confirm* its understanding. A reprompt might be a variant of the same question that was asked before, or it could include asking the user to choose between two tasks that have been assigned similar confidences by the SLU module. For example, in the dialogue in Figure 2 the system utterance in S3 counts as a reprompt because it is a variant of the question in utterance S2.

The features that we extract from the Dialogue Manager are the task-type label, *sys-label*, whose set of values include a value to indicate when the system had insufficient information to decide on a specific task-type, the utterance id within the dialogue (*utt-id*), the name of the prompt played to the user (*prompt*), and whether the type of prompt was a reprompt (*reprompt*), a confirmation (*confirm*), or a subdialogue prompt (a superset of the reprompts and confirmation prompts (*subdial*)). The *sys-label* feature is intended to capture the fact that some tasks may be harder than others. The *utt-id* feature is motivated by the idea that the length of the dialogue may be important, possibly in combination with other features like *sys-label*. The different prompt features for initial prompts, reprompts, confirmation prompts and subdialogue prompts are motivated by results indicating that reprompts and confirmation prompts are frustrating for callers and that callers are likely to hyperarticulate when they have to repeat themselves, which results in ASR errors (Shriberg et al., 1992; Levow, 1998; Walker, Kamm, & Litman, 2000a).

The discourse history features included running tallies for the number of reprompts (*num-reprompts*), number of confirmation prompts (*num-confirms*), and number of subdialogue prompts (*num-subdials*), that had been played before the utterance currently being processed, as well as running percentages (*percent-reprompts, percent-confirms, percent-subdials*). The use of running tallies and percentages is based on previous work suggesting that normalized features are more likely to produce generalized predictors (Litman, Walker, & Kearns, 1999). A feature available for *identifying* problematic dialogues is *dial-duration* that is not available for initial segments of the dialogue.

**Hand Labelling**: As mentioned above, the features obtained via hand-labelling are used to provide a TOPLINE against which to compare the performance of the fully automatic features. The hand-labelled features include human transcripts of each user utterance (*tscript*), a set of semantic labels that are closely related to the system task-type labels (*human-* *label*), age (*age*) and gender (*gender*) of the user, the actual modality of the user utterance (*user-modality*) (one of: nothing, speech, touchtone, speech+touchtone, non-speech), and a cleaned transcript with non-word noise information removed (*clean-tscript*). From these features, we calculated two derived features. The first was the number of words in the cleaned transcript (*cltscript-numwords*), again on the assumption that utterance length is strongly correlated with ASR and SLU errors. The second derived feature was based on calculating whether the *human-label* matches the *sys-label* from the Dialogue Manager (*SLU-success*). This feature is described in detail in the next section.

In the experiments, the features in Figure 7, excluding the hand-labelled features, are referred to as the AUTOMATIC feature set. The experiments test how well misunderstandings can be identified and whether problematic dialogues can be predicted using the AUTOMATIC features. We compare the performance of the AUTOMATIC feature set to the full feature set including the hand-labelled features and to the performance of the AUTOMATIC feature set with and without the *auto-SLU-success* feature. Figure 8 gives an example of the encoding of some of the automatic features for the second exchange of the WIZARD dialogue in Figure 3. The prefix "*e2-*" designates the second exchange. We discuss several of the features values here to ensure that the reader understands the way in which the features are used. In utterance S2 in Figure 3, the system says *Sorry please briefly tell me how I may help you.* In Figure 8, this is encoded by several features. The feature *e2-prompt* gives the name of that prompt, `top-reject-rep`. The feature *e2-reprompt* specifies that S2 is a reprompt, a second attempt by the system to elicit a description of the caller's problem. The feature *e2-confirm* specifies that S2 is not a confirmation prompt. The feature *e2-subdial* specifies that S2 initiates a subdialogue and *e2-num-subdials* encodes that this is the first subdialogue so far, while *e2-percent-subdials* encodes that out of all the system utterances so far, 50% of them initiate subdialogues.

As mentioned earlier, we are also interested in generalizing our problematic dialogue predictor to other systems. Thus, we trained RIPPER using only features that are both automatically acquirable during runtime and independent of the HMIHY task. The subset of features from Figure 7 that fit this qualification are in Figure 9. We refer to them as the AUTO, TASK-INDEPT feature set. Examples of features that are not task-independent include *recog-grammar, sys-label, prompt* and the hand-labelled features.

## 5. Auto-SLU-success Predictor

The goal of the *auto-SLU-success* predictor is to identify, for each exchange, whether or not the system correctly understood the user's utterance. As mentioned above, when the dialogues were transcribed by humans after the data collection was completed, the human labelers not only transcribed the users' utterances, but also labelled each utterance with a semantic category representing the task that the user was asking HMIHY to perform. This label is called the *human-label.* The system's Dialogue Manager decides among several different hypotheses produced by the SLU module, and logs its hypothesis about what task the user was asking HMIHY to perform; the Dialogue Manager's hypothesis is known as the *sys-label.* We distinguish four classes of spoken language understanding outcomes based on comparing the *human-label*, the *sys-label* and recognition results for card and telephone numbers: (1) RCORRECT: SLU correctly identified the task and any digit strings were also

| | | | |
|---|---|---|---|
| *e2-recog*: | can charge no one eight hundred call A T T | | |
| *e2-rg-modality*: | speech-plus-touchtone | *e2-recog-numwords*: | 10 |
| *e2-user-modality*: | speech | *e2-dtmf-flag*: | 0 |
| *e2-rg-grammar*: | Reprompt-gram | *e2-asr-duration*: | 6.68 |
| *e2-top-task* : | dial-for-me | *e2-top-confidence* : | .81 |
| *e2-nexttop-task* : | none | *e2-diff-confidence* : | .81 |
| *e2-salience-coverage*: | 0.000 | *e2-task1* : | 0 |
| *e2-inconsistency*: | 0.000 | *e2-task2*: | 0 |
| *e2-context-shift*: | 0.000 | *e2-task3*: | 0 |
| *e2-prompt* : | top-reject-rep | *e2-task4*: | 0 |
| *e2-reprompt*: | reprompt | *e2-task5*: | 0 |
| *e2-num-reprompts*: | 1 | *e2-task6*: | .81 |
| *e2-percent-reprompts*: | 0.5 | *e2-task7*: | 0 |
| *e2-confirm*: | not-confirm | *e2-task8*: | 0 |
| *e2-num-confirms*: | 0 | *e2-task9*: | 0 |
| *e2-percent-confirms*: | 0 | *e2-task10*: | 0 |
| *e2-subdial*: | subdial | *e2-task11*: | 0 |
| *e2-num-subdials*: | 1 | *e2-task12*: | 0 |
| *e2-percent-subdials*: | 0.5 | *e2-task13*: | 0 |
| *e2-cltscript-numwords*: | 11 | *e2-task14*: | 0 |
| *e2-sys-label*: | DIAL-FOR-ME | *e2-task15*: | 0 |
| *e2-human-label*: | no-info digitstr | *e2-no-info*: | 1 |
| *e2-tscript*: | *epr* I'm trying to call *uh* 1 8 hundred call A T T *nspn* | | |
| *e2-clean-tscript*: | I'm trying to call 1 8 hundred call A T T | | |

Figure 8: Feature encoding for Second Exchange of WIZARD dialogue.

- **Acoustic/ASR Features**
  - recog, recog-numwords, asr-duration, dtmf-flag, rg-modality actual modality of the user utterance.
- **SLU Features**
  - salience coverage, inconsistency, context-shift, top confidence, diff-confidence, auto-SLU-success.
- **Dialogue Manager Features**
  - utterance by utterance: utt-id, reprompt, confirmation, subdial
  - running tallies: num-utts, num-reprompts, percent-reprompts, num-confirms, percent-confirms, num-subdials, percent-subdials, dial-duration correctly recognized; (2) RPARTIAL-MATCH: SLU correctly recognized the task but there was an error in recognizing a calling card number or a phone number; (3) RMISMATCH: SLU did not correctly identify the user's task; (4) NO-RECOG: the recognizer did not get any input to process and so the SLU module did not either. This can arise either because the user did not say anything or because the recognizer was not listening when the user spoke. The rcorrect class accounts for 7481 (36.1%) of the exchanges in the corpus. The rpartial-match accounts for 109 (0.5%) of the exchanges. The rmismatch class accounts for 4197 (20.2%) of the exchanges and the no-recog class accounts for 8943 (43.1%) of the exchanges.

Figure 9: Automatic task-independent features available at runtime.

The *auto-SLU-success* predictor is trained using 45 fully automatic features. These features are the Acoustic/asr features, slu features and Dialogue Manager and Discourse History features, given in Figure 7 . Hand-labelled features were not used.

We evaluate the four-way *auto-SLU-success* classifier by reporting accuracy, precision, recall and the categorization confusion matrix. This classifier is trained on all the features for the whole training set, and then tested on the held-out test set.

Table 1 summarizes the overall accuracy results of the system trained on the whole training set and tested on the test set described in Section 3. The first line of Table 1 represents the accuracy from always guessing the majority class (no-recog); this is the baseline against which the other results should be compared. The second row, labelled automatic, shows the accuracy based on using all the features available from the system modules. This classifier can identify slu errors 47.0% better than the baseline. An experiment was run to see if the cross-validation method described in Section 3 performs worse than using the whole data on the same test set. This experiment showed that there was little loss of accuracy when using cross-validation (0.6%).

| **Features Used** | **Accuracy** |
|---|---|
| baseline (majority class) | 43.1% |
| automatic | 90.1 % |

Table 1: Results for detecting slu Errors using ripper

Figure 10 shows some top performing rules that ripper learns when given all the features. These rules directly reflect the usefulness of the slu features. Note that some of the rules use asr features in combination with slu features such as *salpertime*. Previous studies (Walker et al., 2000c) have also shown slu features to be useful. We had also hypothesized that features from the Dialogue Manager and the discourse history might be useful predictors of slu errors, however these features rarely appear in the rules with the exception of *sys-label*. This is in accordance with previous experiments which show that these features do not add significantly to the performance of the slu only feature set (Walker et al., 2000c).

We also report precision and recall for each category on the held-out test set. The results are shown in Tables 2 and 3. Table 2 shows that the classification accuracy rate is a result of a high rate of correct classification for the rcorrect and no-recog class, at the cost of a lower rate for rmismatch and rpartial-match. This is probably due to the fact that there are fewer examples of these categories in the training set.

In some situations, one might not need to distinguish between the different misunderstanding categories: no-recog, rmismatch and rpartial-match. Therefore, experiments were performed that collapsed these 3 problematic categories into one category (rin-

**if** (sys-label = DIAL-FOR-ME) $\wedge$ (dtmf-flag = 0) $\wedge$ (recog contains "from") $\wedge$ (recog-numwords $\leq$ 8) **then** *rpartial-match*
**if** (sys-label = DIAL-FOR-ME) $\wedge$ (dtmf-flag = 0) $\wedge$ (asr-duration $\geq$ 4.08) $\wedge$ (recog-grammar = Billmethod-gram) $\wedge$ (recog contains "my") $\wedge$ (recog-numwords $\leq$ 8) **then** *rpartial-match*
**if** (spoken-digit = 1) $\wedge$ (salpertime $\leq$ 0.05) $\wedge$ (top-confidence $\leq$ 0.851) **then** *rmismatch*
**if** (spoken-digit = 1) $\wedge$ (salpertime $\leq$ 0.05) $\wedge$ (confpertime $\leq$ 0.076) **then** *rmismatch*
**if** (spoken-digit = 1) $\wedge$ (top-confidence $\leq$ 0.836) **then** *rmismatch*
**if** (spoken-digit = 1) $\wedge$ (salpertime $\leq$ 0.05) $\wedge$ (sys-label = CALLING-CARD) **then** *rmismatch*
**if** (asr-duration $\geq$ 1.04) $\wedge$ (utt-id $\geq$ 2) $\wedge$ (sys-label = DIAL-FOR-ME) $\wedge$ (diff-confidence $\leq$ 0.75) **then** *rmismatch*

Figure 10: A subset of rules learned by *ripper* when given the automatic features for determining *auto-SLU-success*

| Class | Recall | Precision |
|---|---|---|
| RCORRECT | 92.6% | 86.8% |
| NO-RECOG | 98.5% | 97.5% |
| RMISMATCH | 70.6% | 81.0% |
| RPARTIAL-MATCH | 22.7% | 40.0% |

Table 2: Precision and Recall for Test set using Automatic features

CORRECT). This resulted in a recognition accuracy of 92.4%, a 29.4% improvement over the baseline of 63%, which is the percentage of RINCORRECT exchanges. The precision and recall matrix is given in Table 4.

| | RCORRRECT | NO-RECOG | RMISMATCH | RPARTIAL |
|---|---|---|---|---|
| RCORRECT | 2784 | 6 | 211 | 5 |
| NO-RECOG | 9 | 3431 | 44 | 0 |
| RMISMATCH | 409 | 83 | 1204 | 10 |
| RPARTIAL-MATCH | 6 | 0 | 28 | 10 |

Table 3: Confusion Matrix for Test set using Automatic features

| Class | Recall | Precision |
|---|---|---|
| RCORRECT | 91.2% | 89.0 % |
| RINCORRECT | 93.1% | 94.5% |

Table 4: Precision and Recall for Test set using Automatic features

## 6. Problematic Dialogue Predictor

The goal of the PDP is to predict, on the basis of information that it has early in the dialogue, whether or not the system will be able to complete the user's task. The output classes are based on the four dialogue categories described above. However, as HANGUP, WIZARD and TASKFAILURE are treated as equivalently problematic by the system, as illustrated in Figure 5, these 3 categories are collapsed into PROBLEMATIC. Note that this categorization is inherently noisy because it is impossible to know the real reasons why a caller hangs up or a wizard takes over the call. The caller may hang up because she is frustrated with the system, or she may simply dislike automation, or her child may have started crying. Similarly, one wizard may have low confidence in the system's ability to recover from errors and use a conservative approach that results in taking over many calls, while another wizard may be more willing to let the system try to recover. Nevertheless, we take these human actions as a human labelling of these calls as problematic. Given this binary classification, approximately 33% of the calls in the corpus of 4692 dialogues are PROBLEMATIC and 67% are TASKSUCCESS.

### 6.1 Problematic Dialogue Predictor Results

This section presents results for predicting problematic dialogues. Taking into account the fact that a problematic dialogue must be predicted at a point in the dialogue where the system can do something about it, we compare *prediction* accuracy after having seen only the first exchange or the first two exchanges with *identification* accuracy after having seen the whole dialogue. For each of these situations, we also compare results for the AUTOMATIC feature set (as described earlier) with and without the *auto-SLU-success* feature and with the hand-labelled feature *SLU-success*.

Table 5 summarizes the overall accuracy results. The three columns present results for Exchange 1, Exchanges 1&2 and over the whole dialogue. The first row gives the baseline result which represents the prediction accuracy from always guessing the majority class. Since 67.1% of the dialogues are TASKSUCCESS dialogues, we can achieve 67.1% accuracy from simply guessing TASKSUCCESS for each dialogue. The second row gives results using only automatic features, but without the *auto-SLU-success* feature. The third row uses the same automatic features but adds in *auto-SLU-success*. This feature is obtained for both the training and the test set, using the cross-validation method discussed in Section 3. The fourth and fifth rows show results using the subset of features that are both fully automatic and task-independent as described in Section 4.

The automatic results given in row 2 are significantly higher by a paired t-test than the baseline for all three sections of the dialogue (df=866, t=2.1, p=0.035;df=866, t=7.2, p=0.001;df=866 t=13.4, p=0.001).

Rows 6 and 7 show accuracy improvements gained by the addition of hand-labelled features. These rows give a TOPLINE against which to compare the results in rows 2, 3, 4 and 5. Results using all the automatic features plus the hand-labelled *SLU-success* are given in row 6. In these experiments, the hand-labelled *SLU-success* feature is used for training and testing. Comparing this result with the second row shows that if one had a perfect predictor of *auto-SLU-success* in the training and the test set, then this feature would increase accuracy by 5.5% for Exchange 1 (from 70.1% to 75.6%); by 7.6% for Exchanges 1&2 (from 78.1% to 85.7%); and by 5.9% for the whole dialogue (87.0% to 92.9%). These increases are significant by a paired t-test (df=866, t=5.1, p=0.0001; df=866, t=2.1, p=0.035; df=866, t=6.7, p=0.001).

Comparing the result in row 6 with the result in row 3 shows that the *auto-SLU-success* predictor that we have trained can improve performance, but could possibly help more with different training methods. Ideally, the result in row 3, for automatic features plus *auto-SLU-success*, should fall between the figures in rows 2 and 6, and be closer to the results in row 6. With Exchanges 1&2, adding *auto-SLU-success* results in an increase of 1.1% which is not significant (compare rows 2 and 3). For Exchange 1 only, RIPPER does not use the *auto-SLU-success* feature in its ruleset and does not yield an improvement over the system trained only on the automatic features. The system trained on the whole dialogue with automatic features plus *auto-SLU-success* also does not yield an improvement over the system trained without *auto-SLU-success*.

#### 6.1.1 Task-independent Features

Rows 4 and 5 give the results using the AUTO, TASK-INDEPT feature set described in Figure 9 without and with the *auto-SLU-success* feature, respectively. These results are significantly above the baseline using a paired t-test, with Exchanges 1&2 giving an increase of 13.1% (df=866, t=8.6, p=0.001) using TASK-INDEPT features with *auto-SLU-success*. By comparing rows 4 and 5, one observes an increase in the AUTO, TASK-INDEPT features set when the feature *auto-SLU-success* is added using Exchanges 1&2 and whole dialogue. The 1.9% increase for Exchanges 1&2 shows a trend (df=866, t=1.7,p=0.074), whereas the 2% increase for the whole dialogue is statistically significant by a paired t-test (df=866, t=3.0, p=0.003).

Although, the TASK-INDEPT feature sets are a subset of those features used in row 3, it is possible for them to perform better because the TASK-INDEPT features are more general, and because RIPPER uses a greedy algorithm to discover its rule sets. For Exchanges 1&2, the increase from row 3 to row 5 (both of which use *auto-SLU-success*) is not significant. Comparing rows 2 and 4, neither of which use *auto-SLU-success*, one sees a slight degradation in results for the whole dialogue using TASK-INDEPT features. However, the increase from rows 2 to 5 from 78.1% to 80.3% for Exchanges 1&2 is statistically significant (df=866, t=2.0, p=0.042). This shows that using *auto-SLU-success* in combination with the set of TASK-INDEPT features produces a statistically significant increase in accuracy over a set of automatic features that does not include this feature.

| Row | Features | Exchange 1 | Exchange 1&2 | Whole |
|---|---|---|---|---|
| 1 | Baseline | 67.1 | 67.1 | 67.1 |
| 2 | AUTO (no *auto-SLU-success*) | 70.1 | 78.1 | 87.0 |
| 3 | AUTO + *auto-SLU-success* | 69.6 | 79.2 | 84.9 |
| 4 | AUTO, TASK-INDEPT (no *auto-SLU-success*) | 70.1 | 78.4 | 83.4 |
| 5 | AUTO, TASK-INDEPT + *auto-SLU-success* | 69.2 | 80.3 | 85.4 |
| 6 | AUTO + *SLU-success* | 75.6 | 85.7 | 92.9 |
| 7 | ALL (AUTO + Hand-labelled) | 77.1 | 86.9 | 91.7 |

Table 5: Accuracy % results for predicting problematic dialogues.

| Class | Occurred | Predicted | Recall | Precision |
|---|---|---|---|---|
| TASKSUCCESS | 67.0 % | 81.7 % | 88.1 % | 72.5 % |
| PROBLEMATIC | 33.0 % | 18.3 % | 31.6 % | 56.6 % |

Table 6: Precision and Recall with Exchange 1 Automatic Features

The main purpose of these experiments is to determine whether a dialogue is *potentially* problematic, therefore using the whole dialogue is not useful in a dynamic system. Using Exchanges 1&2 produces accurate results and would enable the system to adapt in order to complete the dialogue in the appropriate manner.

#### 6.1.2 Hand-labelled Features

Row 7 in table 5 gives the results using hand-labelled and automatic features including both *SLU-success* and *auto-SLU-success*. By comparing rows 6 and 7, one can see that there is not very much to be gained by adding the other hand-labelled features given in Figure 7 to the hand-labelled and *SLU-success* feature set. Only the increase for Exchange 1 from 75.6% to 77.1% is significant (df=866, t=2.3, p=0.024). For the whole utterance there is actually a degradation of results from 92.9% to 91.7%.

### 6.2 Precision and Recall

The performance of the system that uses automatic features (including *auto-SLU-success*) for the first utterance is given in Table 6. This system has an overall accuracy of 69.6%. These results show that, given the first exchange, the ruleset predicts that 18.3% of the dialogues will be problematic, while 33% of them actually will be. Of the problematic dialogues, it can predict 31.6% of them. Once it predicts that a dialogue will be problematic, it is correct 56.6% of the time.

The performance of the system that uses automatic features for Exchanges 1&2 is summarized in Table 7. These results show that, given the first two exchanges, this ruleset predicts that 20% of the dialogues will be problematic, while 33% of them actually will be. Of the problematic dialogues, it can predict 49.5% of them. Once it predicts that a dialogue will be problematic, it is correct 79.7% of the time. This classifier has an improvement of **if** (e2-salience-coverage $\leq$ 0.7) $\wedge$ (e2-asr-duration $\geq$ 0.04) $\wedge$ (e2-auto-SLU-success = NO-RECOG) **then** *problematic*
**if** (e2-auto-SLU-success = RMISMATCH)$\wedge$ (e2-sys-label = DIAL-FOR-ME) $\wedge$ (e2-asr-duration $\geq$ 3.12) **then** *problematic*
**if** (e1-salience-coverage $\leq$ 0.727) $\wedge$ (e2-salience-coverage $\leq$ 0.706) $\wedge$ (e1-recog contains "help") $\wedge$ (e1-asr-duration $\leq$ 2.44) **then** *problematic*
**if** (e1-top-confidence $\leq$ 0.924) $\wedge$ (e2-auto-SLU-success = RMISMATCH) $\wedge$ (e2-sys-label = CALLING-CARD) **then** *problematic*
**if** (e1-top-confidence $\leq$ 0.924) $\wedge$ (e2-diff-confidence $\leq$ 0.918) $\wedge$ (e2-collect $\leq$ 0.838) $\wedge$ (e2-asr-duration $\geq$ 9.36) $\wedge$ (e2-dial-for-me $\geq$ 0.5) **then** *problematic*

Figure 11: A subset of rules learned by RIPPER when given the automatic features for determining problematic dialogues

**if** (e2-top-confidence $\leq$ 0.897) $\wedge$ (e2-asr-duration $\geq$ 0.04) $\wedge$ (e2-auto-SLU-success = NO-RECOG) **then** *problematic*
**if** (e2-auto-SLU-success = RMISMATCH) $\wedge$ (e2-recog-numwords $\leq$ 7) $\wedge$ (e2-asr-duration $\geq$ 2.28) **then** *problematic*
**if** (e2-salience-coverage $\leq$ 0.9) $\wedge$ (e2-asr-duration $\geq$ 0.04) $\wedge$ (e2-inconsistency $\geq$ 0.022) $\wedge$ (e1-asr-duration $\leq$ 3.96) $\wedge$ (e2-inconsistency $\geq$ 0.18) **then** *problematic*
**if** (e1-salience-coverage $\leq$ 0.667) $\wedge$ (e2-salience-coverage $\leq$ 0.692) $\wedge$ (e1-recog contains "help") $\wedge$ (e1-asr-duration $\leq$ 7.36) **then** *problematic*
**if** (e1-top-confidence $\leq$ 0.924) $\wedge$ (e2-diff-confidence $\leq$ 0.918) $\wedge$ (e2-recog contains "my") $\wedge$ (e1-asr-duration $\geq$ 4) **then** *problematic*
**if** (e1-salience-coverage $\leq$ 0.647) $\wedge$ (e1-asr-duration $\geq$ 10.24) $\wedge$ (e2-asr-duration $\geq$ 5.32) **then** *problematic*

Figure 12: A subset of rules learned by RIPPER when given the TASK-INDEPT features for determining problematic dialogues

17.87% in recall and 23.09% in precision, for an overall improvement in accuracy of 9.6% over using the first exchange alone.

### 6.3 Examination of the Rulesets

A subset of the rules from the system that uses automatic features for Exchanges 1&2 are given in Figure 11 (row 3, table 5). One observation from these hypotheses is the classifier's preference for the *asr-duration* feature over the feature for the number of words recognized (*recog-numwords*). One would expect longer utterances to be more difficult, but the learned rulesets indicate that duration is a better measure of utterance length than the number of words. Another observation is the usefulness of the SLU confidence scores and the SLU salience-coverage in predicting problematic dialogues. These features seem to provide good general indicators of the system's success in recognition and understanding. The fact that the main focus of the rules is detecting ASR and SLU errors and that none of the DM behaviors are used as predictors also indicates that, in all likelihood, the DM is performing as well as it can, given the noisy input that it is getting from ASR and SLU. An alternative view is that two utterances are not enough to provide meaningful dialogue features such as counts and percentages of reprompts, confirmations, etc..

| Class | Occurred | Predicted | Recall | Precision |
|---|---|---|---|---|
| TASKSUCCESS | 67.0 % | 80.0 % | 94.8 % | 79.1 % |
| PROBLEMATIC | 33.0 % | 20.0 % | 49.5 % | 79.7 % |

Table 7: Precision and Recall with Exchange 1&2 Automatic Features

One can see that the top two rules use *auto-SLU-success*. The first rule basically states that if there is no recognition for the second exchange (as predicted by the *auto-SLU-success*) then the dialogue will fail. The second rule is more interesting as it states if a misunderstanding has been predicted for the second exchange and the system label is DIAL-FOR-ME and the utterance is long then the system will fail. In other words, the system frequently misinterprets long utterances as DIAL-FOR-ME resulting in task failure.

Figure 12 gives a subset of the ruleset for the TASK-INDEPT feature set for Exchanges 1&2. One can see a similarity between this ruleset and the one given in Figure 11. This is due to the fact that when all the automatic features are available, RIPPER has a tendency to pick out the more general task-independent ones, with the exception of *sys-label*. If one compares the second rule in both figures, one can see that RIPPER uses *recog-numwords* as a substitute for the task-specific feature *sys-label*.

### 6.4 Cross-validation Method vs. Hand-labelled-training Method

As mentioned above, an alternative to training the PDP on the automatically derived *auto-SLU-success* feature is to train it on the hand-labelled *SLU-success* while still testing it on the automatic feature. This second method is referred to as "hand-labelled-training" and the resulting feature is *hlt-SLU-success*. This may provide a more accurate model but it may not capture the characteristics of the automatic feature in the test set. Table 8 gives results for the two methods. One can see from this table that there is a slight, insignificant increase in accuracy for Exchange 1 and the whole dialogue using the hand-labelled-training method. However, the totally automated method yields a better result (79.2% compared to 77.4%) for Exchanges 1&2, which as mentioned above, is the most important result for these experiments. This increase shows a trend but is not significant (df=866, t=1.8, p=0.066). The final row of the table gives the results using the hand-labelled feature *SLU-success* in both the training and testing and is taken as the topline result.

| Features | Exchange 1 | Exchange 1&2 | Whole |
|---|---|---|---|
| Baseline | 67.1 | 67.1 | 67.1 |
| AUTO | 70.1 | 78.1 | 87.0 |
| AUTO + *hlt-SLU-success* | 70.4 | 77.4 | 86.2 |
| AUTO + *auto-SLU-success* | 69.6 | 79.2 | 84.9 |
| AUTO + *SLU-success* | 75.6 | 85.7 | 92.9 |

Table 8: Accuracy % results including *hlt-SLU-success* derived using the hand-labelled-training method

| Features | Exchange 1 | Exchange 1&2 | Whole |
|---|---|---|---|
| Baseline | 67.1 | 67.1 | 67.1 |
| ASR | 66.7 | 75.9 | 85.6 |
| SLU | 67.7 | 71.9 | 79.8 |
| Dialogue | 65.5 | 74.5 | 82.6 |
| Hand-labelled | 76.9 | 84.7 | 86.2 |
| *Auto-SLU-success* | 69.0 | 70.9 | 77.1 |
| *Hlt-SLU-success* | 69.0 | 74.1 | 77.2 |

Table 9: Accuracy % results for subsets of features

### 6.5 Feature Sets

It is interesting to examine what types of features are the most discriminatory in determining whether a dialogue is problematic or not. RIPPER was trained separately on sets of features based on the groups given in Figure 7, namely Acoustic/ASR, SLU, Dialogue and Hand-labelled (including *SLU-success*). These results are given in Table 9.

For Exchange 1, only the SLU features, out of the automatic feature sets, yields an improvement over the baseline. Interestingly, training the system on the ASR yields the best result out of the automatic feature sets for Exchange 1&2 and the whole dialogue. These systems, for example, use *asr-duration*, number of recognized words, and type of recognition grammar as features in their ruleset.

Finally, we give results for the system trained only on *auto-SLU-success* and *hlt-SLU-success*. One can see that there is not much difference in the two sets of results. For Exchanges 1&2, the system trained on *hlt-SLU-success* has an accuracy which is significantly higher than the system trained on *auto-SLU-success* by a paired t-test (df=866, t=3.0, p=0.03). On examining the ruleset, one finds that the *hlt-SLU-success* uses RPARTIAL-MISMATCH where the *auto-SLU-success* ruleset does not. The lower accuracy may be due to the fact that the *auto-SLU-success* predictor has a low recall and precision for RPARTIAL-MISMATCH as seen in Table 2.

### 6.6 Types of Problematic Dialogues

As mentioned in Section 2, there are 3 types of problematic dialogues: TASKFAILURE, WIZARD and HANGUP. In order to determine whether some of these types of problematic dialogues are more difficult to predict than others, we conducted a post-hoc analysis of the proportion of prediction failures for each type of problematic dialogue. Since we were primarily interested in the performance of the PDP using the full automatic feature set, after having seen Exchanges 1&2, we conducted our analysis on this version of the PDP. Table 10 shows the distribution of the 4 types of dialogue in the test set and whether the Exchanges 1&2 PDP was able to predict correctly that the dialogue would be TASKSUCCESS or PROBLEMATIC. One can see that the worst performing category is TASKFAILURE and that the PDP predicts incorrectly that 68.5% of the TASKFAILURE dialogues are TASKSUCCESS.

| True Values | Predicted successful | Predicted problematic | total |
|---|---|---|---|
| TASKSUCCESS | 94.1% (548) | 5.9% (34) | 67.1% (582) |
| TASKFAILURE | 68.5% (74) | 31.5% (34) | 12.0% (108) |
| WIZARD | 41.3% (43) | 58.7% (61) | 12.5% (104) |
| HANGUP | 35.6% (26) | 64.4% (47) | 8.4% (73) |
| Total | 79.7% (691) | 20.3% (176) | 100% (867) |

Table 10: Matrix of recognized TASKSUCCESS and TASKFAILURES

| True Values | Predicted successful | Predicted problematic | Total |
|---|---|---|---|
| TASKSUCCESS | 61.1% (66) | 38.9% (42) | 50% (108) |
| TASKFAILURE | 21.3% (23) | 78.6% (85) | 50% (108) |
| Total | 41.2% (89) | 58.8% (127) | 100% (216) |

Table 11: Matrix of recognized TASKSUCCESS and TASKFAILURES using equal training and testing

One reason that this might occur is that this sub-category of dialogues are much more difficult to predict since in this case the HMIHY system has no indication that it is not succeeding in the task. However, another possibility is that the PDP performs poorly on this category because there are fewer examples in the training set, although it does better on the HANGUP subset, which is about the same proportion. We can eliminate the first possibility by examining how a learner performs when trained on equal proportions of TASKSUCCESS and TASKFAILURE dialogues. We conducted an experiment using a subset of TASKSUCCESS dialogues in the same proportion as TASKFAILURE for the training and the test set and trained a second PDP using the fully automatic Exchange 1&2 features. This resulted in a training set of 690 dialogues and a test set of 216. The binary classifier has an accuracy of 70%, the corresponding recognition matrix is presented in table 11. The results show that fewer TASKFAILURES are predicted as successful, suggesting that TASKFAILURES are not inherently more difficult to predict than other classes of problematic dialogues. Below we discuss the potential of using RIPPER's loss ratio to weight different types of classification errors in future work.

## 7. Related Work

The research reported here is the first that we know of to automatically analyze a corpus of logs from a spoken dialogue system for the purpose of learning to *predict* problematic situations. This work builds on two strands of earlier research. First, this approach was inspired by work on the PARADISE evaluation framework for spoken dialogue systems which utilizes both multivariate linear regression and CART to predict user satisfaction as a function of a number of other metrics (Walker, Litman, Kamm, & Abella, 1997; Walker et al., 2000a). Research using PARADISE has found that task completion is always a major predictor of user satisfaction, and has examined predictors of task completion. Here, our goals are similar in that we attempt to understand the factors that predict task completion. Secondly, this work builds on earlier research on learning to *identify* dialogues in which the user experienced poor speech recognizer performance (Litman et al., 1999). Because that work was based on features synthesized over the entire dialogue, the hypotheses that were learned could not be used for prediction during runtime. In addition, in contrast to the current study, the previous work automatically approximated the notion of a good or bad dialogue using a threshold on the percentage of recognition errors. There is a danger of this approach being circular when recognition performance at the utterance level is a primary predictor of a good or bad dialogue. In this work, the notion of a good (TASKSUCCESS) and bad (PROBLEMATIC) dialogue was labelled by humans.

In previous work, (Walker et al., 2000b) reported results from training a problematic dialogue predictor in which they noted the extent to which the hand-labelled *SLU-success* feature improves classifier performance. As a result of this prior analysis, in this work we report results from training an *auto-SLU-success* classifier for each exchange and using its predictions as an input feature to the Problematic Dialogue Predictor. There are a number of previous studies on predicting recognition errors and user corrections which are related to the *auto-SLU-success* predictor that we report on here (Hirschberg et al., 1999; Hirschberg, Litman, & Swerts, 2000, 2001b; Levow, 1998; Litman, Hirschberg, & Swerts, 2000; Swerts, Litman, & Hirschberg, 2000).

(Hirschberg et al., 1999) apply RIPPER to predict recognition errors in a corpus of 2067 utterances. In contrast to our work, they utilize prosodic features in combination with acoustic confidence scores. They report a best-classifier accuracy of 89%, which is a 14% improvement over their baseline of 74%. This result can be compared with our binary *auto-SLU-success* predictor (RCORRECT vs. RINCORRECT) discussed in Section 5. Examination of the rules learned by their classifier suggests that durational features are important. While we do not use amplitude or F0 features, we do have an *asr-duration* feature which is logged by the recognizer. Without any of the other prosodic features, the *auto-SLU-success* predictor has an accuracy of 92.4%, a 29.4% improvement over the baseline of 63%. It is possible that including prosodic features in the *auto-SLU-success* predictor could improve this result even further.

Previous studies on error correction recognition are also related to our method of misunderstanding recognition. (Levow, 1998) applied similar techniques to learn to distinguish between utterances in which the user originally provided some information to the system, and *corrections*, which provided the same information a second time, following a misunderstanding. This may be more related to our research than it first appears since corrections are often misunderstood due to hyper-articulation. Levow's experiments train a decision tree using features such as duration, tempo, pitch, amplitude, and within-utterance pauses. Examination of the trained tree in this study also reveals that the durational features are the most discriminatory. Similarly in our experiments, RIPPER uses *asr-duration* frequently in the developed rule set. Levow obtains an accuracy rate of 75% with a baseline of 50%.

(Swerts et al., 2000) and (Hirschberg et al., 2001b) perform similar studies for automatically identifying corrections using prosody, ASR features and dialogue context. Corrections are likely to be misrecognized, due to hyperarticulation. They observe that corrections that are more distant from the error they correct, are more likely to exhibit prosodic differences. Their system automatically differentiates corrections from non-corrections with an error rate of 15.72%. Dialogue context is used in the study by (Hirschberg, Litman, & Swerts, 2001a), whereby they incorporate whether the user is aware of a mistake at the current utterance to help predict misunderstandings and misrecognition of the previous utterances. This study is similar to ours in that they use a predicted feature about an utterance (the 'aware' feature) to predict concept or word accuracy, as we use a predicted feature *auto-SLU-success* in the PDP. However, our *auto-SLU-success* feature is automatically available at the time the prediction is being made, whereas they are making the predictions retroactively. In addition, they train their system on the hand-labelled feature rather than the predicted one which they leave as further work.

(Kirchhoff, 2001) performs error correction identification using task independent acoustic and discourse variables. This is a two way distinction between positive and negative error correction. She uses two cascaded classifiers, the first is a decision tree trained using 80% of the data and validating on 10%. Examples that have confidence scores below a threshold go into an exception training set for a second classifier. During testing, if confidence scores are below a threshold then the utterance is passed onto the second classifier. She finds that the most discriminatory features are dialogue context (the type of previous system utterance) followed by lexical features, with prosodic features being the least discriminatory. The system recognizes error corrections with an accuracy of 90% compared to a baseline of 81.9%. In this study (Kirchhoff, 2001) deliberately eschews the use of system specific features, while in our work, we examine the separate contribution of different feature sets. Our results suggest that the use of more general features does not negatively impact performance.

(Krahmer, Swerts, Theune, & Weegels, 1999a) and (Krahmer, Swerts, Theune, & Weegels, 1999b) look at different features related to responses to problematic system turns. The *disconfirmations* they discuss are responses to explicit or implicit system verification questions. They observe that disconfirmations are longer, have a marked word order, and contain specific lexicon such as "no". In addition, there are specific prosodic cues such as boundary tones and pauses. Some of these features such as length, choice of words are captured in our RIPPER ruleset as discussed above.

As described in Section 5, two methodologies were compared for incorporating the feature *SLU-success* into the PDP. The first was to use the hand-labelled feature in the training set, the second to perform separate experiments to predict the feature for the training set. As the features in the training set are automatically predicted, it is hoped that the system would pick up the idiosyncrasies of the noisy data. This training method has been used previously in (Wright, 2000) where automatically identified intonation event features are used to train an automatic speech-act detector. These automatically derived features provide a better training model than the hand-labelled ones. This is true also in the current study as discussed in Section 6.1.

## 8. Discussion and Future Work

This paper reports results on automatically training a Problematic Dialogue Predictor to predict problematic human-computer dialogues using a corpus of 4692 dialogues collected with the *How May I Help You* spoken dialogue system. The Problematic Dialogue Predictor can be immediately applied to the system's decision of whether to transfer the call to a human customer care agent, or be used as a cue to the system's Dialogue Manager to modify its behavior to repair the problems identified. The results show that: (1) Most feature sets significantly improve over the baseline; (2) Using automatic features from the whole dialogue, we can identify problematic dialogues 20% better than the baseline; (3) Just the first exchange provides significantly better prediction (3%) than the baseline; (4) The second exchange provides an additional significant (13%) improvement, (5) A classifier based on task-independent automatic features performs slightly better than one trained on the full automatic feature set.

The improved ability to predict problematic dialogues is important for fielding the HMIHY system without the need for the oversight of a human customer care agent. These results are promising and we expect to be able to improve upon them, possibly by incorporating prosody into the feature set (Hirschberg et al., 1999) or expanding on the SLU feature sets. In addition, the results suggest that the current PDP is likely to generalize to other dialogue systems.

In future work, we plan to integrate the learned rulesets into the HMIHY dialogue system and evaluate the impact that this would have on the system's overall performance. There are several ways we might be able to show this. Remember that one use of the PDP is to improve the system's decision of whether and when to transfer a call to the human customer care agent. The other use would be as input to the Dialogue Manager's dialogue strategy selection mechanism. Demonstrating the utility of the PDP for dialogue strategy selection requires experiments that test out several different ways that this information could be used by the Dialogue Manager. Demonstrating the utility of the PDP on the decision to transfer a call necessarily involves examining the tradeoffs among different kinds of errors. This is because every call that the HMIHY system can handle successfully saves a company the cost of using a human customer care agent to handle the call. Thus, we can associate this cost with the decision that HMIHY makes to transfer the call. When HMIHY transfers the call unnecessarily, we call this cost the *lost automation cost*. On the other hand, every call that HMIHY attempts to handle and fails, would potentially accrue a different cost, namely the lost revenue from customers who become irritated with faulty customer service and take their business elsewhere. We call this cost the *system failure cost*. In the results that we presented here, we report only overall accuracy results and treat *lost automation cost* and *system failure cost* as equally costly. However, in any particular installation of the HMIHY system, there may be differences between these costs that would need to be accounted for in the training of the PDP. It would be possible to use RIPPER to do this, if these costs were known, by using its ability to vary the loss ratio.

Another potential issue for future work is the utility of a dialogue level predictor, e.g. the PDP, vs. an utterance level predictor, e.g. the *auto-SLU-success* predictor, for the goal of automatically adapting a system's dialogue strategy. This is shown to be effective in (Litman & Pan, 2000), where they use a problematic dialogue detector in order to adapt the dialogue strategy for a train enquiry system. It would be possible, and others have argued (Levow, 1998; Hirschberg et al., 1999; Kirchhoff, 2001) that the dialogue manager's adaptation decisions can be made on the basis of local behavior, i.e. on the basis of recognizing that the current utterance has been misunderstood, or that the current utterance is a correction. However, it is clear that the decision to transfer the call to a human customer care agent cannot be made on the basis of only local information because the system can often recover from a single error. Thus, we expect that the ability to be able to predict the dialogue outcome as we do here will continue to be important even in systems that use local predictors for understanding and correction.

## 9. Acknowledgments

Thanks to Ron Prass, Diane Litman, Richard Sutton, Mazin Rahim and Michael Kearns for discussions on various aspects of this work.